\documentclass{article}

\usepackage{Resources/PRIMEarxiv}

\usepackage[utf8]{inputenc} 
\usepackage[T1]{fontenc}    
\usepackage{hyperref}       
\usepackage{url}            
\usepackage{booktabs}       
\usepackage{amsfonts}       
\usepackage{nicefrac}       
\usepackage{microtype}      
\usepackage{lipsum}
\usepackage{soul,xcolor}
\usepackage{blkarray}

\usepackage{caption}
\usepackage{subcaption}
\usepackage{float}
\usepackage{amsmath}
\usepackage{fancyhdr}       
\usepackage{graphicx}       
\graphicspath{{media/}}     

\title{Spiking Neural Operators for Scientific Machine Learning
}

\author{%
  Adar~Kahana, Qian~Zhang, Leonard~Gleyzer, George Em~Karniadakis\\
  Division of Applied Mathematics\\
  Brown University\\
  Providence, RI 02912 \\
  \texttt{\{adar\_kahana, qian\_zhang1, leonard\_gleyzer, george\_karniadakis\}@brown.edu} \\
}

\begin{document}
\maketitle

\begin{abstract}
  The main computational task of Scientific Machine Learning (SciML) is function regression, required both for inputs as well as outputs of a simulation. Physics-Informed Neural Networks (PINNs) and neural operators (such as DeepONet) have been very effective in solving Partial Differential Equations (PDEs), but they tax computational resources heavily and cannot be readily adopted for edge computing. Here, we address this issue by considering Spiking Neural Networks (SNNs), which have shown promise in reducing energy consumption by two orders of magnitude or more. We present a SNN-based method to perform regression, which has been a challenge due to the inherent difficulty in representing a function's input domain and continuous output values as spikes. We first propose a new method for encoding continuous values into spikes based on a triangular matrix in space and time, and demonstrate its better performance compared to the existing methods. Next, we demonstrate that using a simple SNN architecture consisting of Leaky Integrate and Fire (LIF) activation and two dense layers, we can achieve relatively accurate function regression results. Moreover, we can replace the LIF with a trained Multi-Layer Perceptron (MLP) network and obtain comparable results but three times faster. Then, we introduce the DeepONet, consisting of a branch (typically a Fully-connected Neural Network, FNN) for inputs and a trunk (also a FNN) for outputs. We can build a spiking DeepONet by either replacing the branch or the trunk by a SNN. We demonstrate this new approach for classification using the SNN in the branch, achieving results comparable to the literature. Finally, we design a spiking DeepONet for regression by replacing its trunk with a SNN, and achieve good accuracy for approximating functions as well as inferring solutions of differential equations.
\end{abstract}


\section{Introduction}


Over the past few years there has been a growing interest in the applied mathematics community for using Machine Learning (ML) based algorithms for approximating functions, solving differential equations, representing operators, and so on \cite{PANG2019270,pmlr-v125-jin20a,pinns_heat,MAO2020112789,RAISSI2019686,lu2021learning}. Known for their ability to fit non-linear problems, ML algorithms and specifically Deep Learning (DL) algorithms have been widely used for solving problems that seemed impossible before. Some applications use ML for large-scale problems, where training takes long but inference is very fast compared to existing analytical methods \cite{briggs,brown1992implications}. Other applications involve solving very complex problems by fitting observations  instead of directly computing a complex expression \cite{lulu_bifidelity}.

A breakthrough in the field of solving PDEs was introduced by Raissi \textit{et al.} \cite{RAISSI2019686}. They proposed a PINN, that receives as input data of the solution of the PDE (domain points and corresponding solution function values), and fits the solution on the entire domain, not requiring mesh generation. PINNs have shown great results for many applications, and have been empirically shown to be robust and efficient \cite{mao2020physics,pang2019fpinns,YIN2021113603,cai2022physics,ZHANG2022Analyses}.

A drawback of DL based methods in general, PINNs included, is that they require large computational effort for training. For example, to train a PINN to detect a surface crack using ultrasound data takes 9 hours on a V100 nVidia GPU \cite{shukla2020physics}. We pursue a lighter method for function regression that supports parallelism and acceleration, while staying as robust as the existing methods. This paper is a first step in that direction.

\subsection{Biologically-plausible neural networks}

The brain is often regarded as the most efficient learning machine in existence. Comparing its energy consumption with that of a high-performance computer has shown that the brain uses many orders of magnitude less power \cite{mcdonnell2014engineering}. Despite its low power usage, the brain is able to maintain high accuracy across a variety of multiple tasks (e.g., signal processing/interpretation, decision making, motor control, etc.). An interesting challenge is to create biologically-plausible algorithms by implementing components inspired by parts of the brain, e.g. the membrane, synapse, etc., \cite{lapique1907recherches,cronin1987mathematical,guo2020self}, as well as to utilize existing mathematical models for neural components \cite{song2000competitive,van2000stable} in order to create a learning algorithm that can approach the effectiveness of the human brain.

In this paper we discuss algorithms from the software side. We note that there is ongoing research on this topic on the hardware side. Researchers are actively developing special purpose computers called neuromorphic computers, which contain electronic analog and/or digital circuits that mimic biological neurons in terms of architecture and activity. A very recent example of such a neuromorphic chip is the Intel Loihi 2 core processing unit \cite{rueckauer2022nxtf}, which also offers the Lava software framework -- a simulator for running neuromorphic programs before integrating them with the hardware.

\subsection{Operator regression - DeepONets}

DeepONets, introduced by Lu \textit{et al}. \cite{lu2021learning}, are neural polymorphic networks used to approximate mathematical operators. The mathematical theory behind DeepONets relies on the universal approximation theorem for operators (a mapping between function spaces) \cite{chen1995universal}. DeepONets have two major architectural components: the branch net and the trunk net, both of which are artificial neural networks (ANNs) (can be convolutional, fully-connected, recurrent, etc.). The branch takes in a representation of a function from the input function space, while the trunk takes in a point in the domain of the output function. Together, the output of a DeepONet is the evaluation of the output function of an operator (for a given input function) at a point in the output function's domain. Mathematically, DeepONet can be formulated as $G(y)(x)=\sum{\alpha_i(y) \phi_i(x)}$, where $\alpha_i(y)$ is the branch network (which maps the function space), and $\phi_i(x)$ is the trunk network, which can be interpreted as a set of basis functions.

DeepONets have demonstrated high accuracy in approximating operators, including complex ones such as fractional operators, stochastic operators and more \cite{lu2021learning,lin2021operator}. However, compared with standard numerical methods, DeepONets do not match in terms of accuracy, usually reaching a minimum of around $10^{-5}$ (using standard $l_2$-norm to compare the predicted solution to an analytic solution), while numerical methods may reach even machine precision. That said, after training, DeepONets can perform rapid inference and do not need to solve the problem from scratch. This is inferior accuracy to the highly accurate methods that take hours to solve, however DeepONets can infer a solution in a fraction of a second. This brings many opportunities, especially in trying to approximate brain-like operators with unknown formulation in a highly efficient implementation.

\subsection{Spikes in the brain}

In the human brain, data is transmitted between neurons through biological components called synapses (discussed in section \ref{synapse}). The inputs to a neuron are electrical potentials. The flow of information in the brain happens according to spike trains, and based on the transmission details of these spikes between neurons, the brain makes decisions and reacts to stimuli. A crucial component of this process is the membrane, which is a cellular gate for incoming potentials entering the neuron. The input voltage to the neuron is controlled by the membrane and if a certain voltage threshold is crossed, the neuron ``fires'' an action potential, or in other words, generates a spike. This spike is then transferred to the subsequent neurons through the synapse.

Standard ANNs, while having some similarities to the brain, are mostly biologically implausible. Many advanced architectures allow for learning different input shapes (such as images, audio signals, tabular data, etc.) and also different input data types (decimal numbers, complex numbers, integer encoding, etc.). This, among other things, differs ANNs from the brain, making them less biologically plausible. In addition, the spikes train in the brain propagates through time, whereas in ANNs the input data does not necessarily have time dependency. Furthermore, ANNs rely on back-propagation \cite{rumelhart1986learning} in order to adjust the network parameters, while back-propagation in the brain is thought to be highly implausible \cite{lillicrap2020backpropagation}. 

\begin{figure}
    \centering
    \includegraphics[width=15cm]{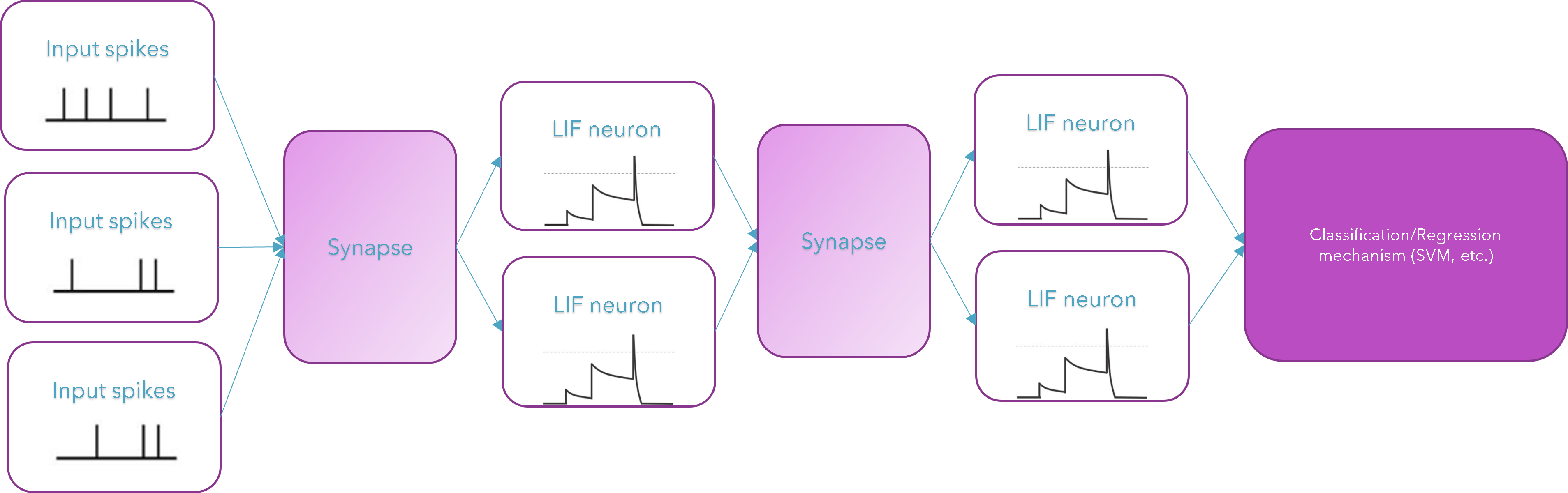}
    \caption{A typical sketch of a SNN architecture. The spiking data is used as inputs to LIF neurons (membranes), and then fed into a synapse. In this architecture, we have a membrane followed by a synapse and another membrane followed by a second synapse. Finally, a classification of regression mechanism should be applied. The gradient color reflects a trainable layer, and the solid color reflects the classification or regression mechanism that could be implemented using SVM or any other (not necessarily deep learning based) method.}
    \label{fig:snn_general}
\end{figure}

To create a learning algorithm that handles spike trains (binary, time dependent input) we adopt the concept of Spiking Neural Networks (SNNs).
SNN is a general terminology for a neural network that handles spiking data. The input to the network is a set of time dependent, usually binary data samples. A typical sketch of a SNN is shown in figure \ref{fig:snn_general}, consisting of input spikes, two membranes, two synapses, and a classification or regression mechanism at the end. Note that a non-differentiable operation happens between the first synapse and the second synapse, creating a challenge for training the SNN. In addition, depending on the desired output, the classification or regression mechanism should be carefully chosen.

There are three dominant methodologies for training a SNN:

\begin{itemize}
    \item \textbf{ANN-SNN conversion}: Conversion methods usually involve training an ANN and integrating it into a spiking framework. The advantages are competitive accuracy and flexibility in terms of network architecture (since the ANN can be state-of-the-art method for the task) but the main drawback is that training the ANN is still costly.
    \item \textbf{Surrogate models}: Surrogate models try to do back-propagation in the SNN, which is complicated since working with spikes involves thresholding and other non-differentiable procedures. The surrogate models compute the approximated gradients in the back-propagation. Therefore, one disadvantage is that the accuracy depends on the approximation of the gradients (which is also a costly procedure). In addition, only specific functions have surrogate models and not every architecture supports that.
    \item \textbf{Pure SNN}: By this we mean creating a SNN with biologically plausible components only and avoid back propagation.
\end{itemize}

As demonstrated by Diehl \textit{et al.} \cite{diehl2015unsupervised}, SNNs can be used for efficient learning, and can be implemented in a robust fashion. In their work the authors proposed a spiking convolutional neural network that achieves high accuracy relative to the other SNN benchmark methods. The benchmark data used was MNIST digit recognition dataset, which is a favourable benchmark for SNNs since it is easy to spike-encode and classify. Many researchers that work on SNNs naturally prefer classification tasks \cite{lee2016training,tavanaei2019deep,sengupta2019going,kour2014real}. In this paper, we show that regression can also be done in this paradigm.

\subsection{Learning models for SNNs}

To train a SNN we implement components that mimic the behavior of different elements in the human brain. One of them is the membrane, mentioned in the previous subsection. There are several mathematical models that simulate the membrane behavior. A popular one is the Leaky Integrate and Fire (LIF), introduced by Louis Lapicque in 1907 \cite{lapique1907recherches}. In this model, the membrane dynamics are defined by a Resistor-Capacitor (RC) circuit. This can then be formulated using an Ordinary Differential Equation (ODE) and solved using many numerical approximation schemes, including forward Euler which is widely used for this task. The LIF model is very popular mainly since it is easy to understand and implement efficiently.

Another model widely used in the neuroscience community is the Hodgkin-Huxley membrane model \cite{hodgkin1952currents}. This model takes into account three different types of ion currents - potassium current, sodium current and current leak. These lead to a more complex yet more accurate model of the membrane voltage at given time. This model can capture spiking behavior better when the spikes are not perfectly binary. There are several other membrane models such as adaptive LIF, fractional LIF, stochastic membrane models, and many more \cite{gerstner2014neuronal,lundstrom2008fractional,stein1965theoretical}. The trade-off when choosing a model usually is between accuracy and efficiency.

Another important component in a SNN is the synapse. The synapse is the component where the learning process happens. In the brain, the synapse is in charge of transmitting spikes between neurons. In other words, the synapses in the brain are the connections between neurons, which enable transmission of information (in this case, spike trains). Since the spike trains are based on time, the synapses in the brain can be thought of as sequential, and if we focus on a single synapse we can say it connects a set of pre-synaptic neurons to a set of post-synaptic neurons. The synapse, in general, determines the connection between each and every pair of neurons.

One way to implement a synapse is using a learning rule called Spike Timing Dependent Plasticity (STDP) \cite{van2000stable}. This rule has been designed to mimic plasticity in the brain. STDP, simply put, means that if a pre-synaptic neuron fired before a post-synaptic neuron, the connection between them is increased. Otherwise it is decreased (also known as potentiation and depression in the synapse). The quantity with which the synapse updates depends on the difference in time between the spikes. For example, if the pre-synaptic neuron spiked shortly before the post-synaptic neuron, the connection between them is largely increased, but if the pre-synaptic spike happened long time before, the connections is slightly increased. The implementation of the STDP rule is usually very costly, especially when the network has several layers (several synapses) and each layer has many input neurons. The most influential factor on the efficiency is the number of total spikes in the system, which dictates the bandwidth of information that flows through the network. Moreover, since the spikes are binary, invoking the sparsity argument, more spikes mean less efficiency.

A popular method for implementing a synapse is using a fully-connected (dense) layer, which is well known in ANNs. Instead of applying the learning rule, the weights of the synapse are updated by a gradient descent algorithm. In this fashion, back-propagation is used and the network is less biologically plausible. On the other hand, with the efficient Stochastic Gradient Descent (SGD) methods developed through the years for ANNs, using a dense layer may result in much faster training times and also relatively high accuracy compared to STDP, but not competitive with state-of-the-art methods in the field of classification, such as \cite{an2020ensemble}. Many authors use this method \cite{eshraghian2021training,kim2020revisiting,kulkarni2018spiking,stimberg2019brian}. The implementation of a synapse is costly compared to a dense layer, as demonstrated later in this paper. In addition, many authors suggest using a dense layer as the last hidden layer (only the last one), and train this layer for the classification, or as discussed in this paper for regression. In this case back-propagation is less expensive, and problems like vanishing gradients \cite{hochreiter1998vanishing} are avoided.

Other algorithms have been proposed to help SNNs achieve the accuracy and efficiency of ANNs (compromising on biological plausibility). One of which is called Batch Normalization Through Time (BNTT), proposed by Kim \textit{et al.} \cite{kim2020revisiting}. The BNTT algorithms suggests splitting the spiking inputs into mini-batches where all samples in the mini-batch are from the same time step (further explanation about time steps and spikes train is given in section \ref{snnExplained}). After a train step, the batch is normalized according to the mean and standard deviation of the samples in the batch. Then, a set of trainable weights multiply the normalized data. Global mean and standard deviation are kept for inference. The authors present interesting behavior of the weights that multiply the normalized data, showing Gaussian patterns that emphasize different times. Also, they show that the majority of spikes occur before a certain time step, enabling them to reduce the number of time steps and thus the complexity of the network. This is an innovative idea of how to implement a well-known method popular for ANNs but in the SNN context.

There are not many studies of SNNs supported by fundamental mathematical theorems. One paper by Maass \cite{maass1997fast}, proposes a universal approximation theorem for spikes.
This theorem shows that every function can be approximated by a sufficiently deep noisy spiking network. Noisy spikes have spiking behavior (sharp jumps and drops), but are not completely binary. They are characterized by random distributions around 1 and 0, 1 when a spike occurs and 0 when the neuron is idle. In the current work we do not consider noisy spikes.

\section{Spiking neural networks}\label{snnExplained}

In this section we discuss in detail how the different components of the SNN are designed. We begin with methods to encode input data into spiking data. We show how the popular methods for encoding data into spikes in classification tasks can be used for regression as well, and propose a new method specifically for regression. Then we present the leaky integrate-and-fire model used to model a membrane and compute output spikes based on the membrane trace. Finally, we present the learning part, which is performed in the synapse by a learning rule.

\subsection{Spike encoding}\label{encoding}

The first step of creating a SNN is handling input data. The type of the input data for ANNs is usually decimal numbers (pixels in images, points on an interval, etc.), while the SNN expects binary inputs. Therefore, we need to encode the input data into spikes. We use the terminology \textit{spiking data} to distinguish between encoded data and raw (non-spiking) data. There are several methods for encoding data into spikes. For example, Kim \textit{et al.} \cite{kim2022rate} showed that using a fully-connected method to encode data into spikes yields better results for the SNN training than the popular rate encoding, discussed in the next subsection. Another important realization from their work is the impact of different encoding methods on the performance of the SNNs. Since there is no optimal method to encode data into spikes, this is an interesting and high impact subject to explore. The results using the different encoding methods we explore in this work are discussed in section \ref{results}.

\subsubsection{Rate encoding}

Rate encoding is the most popular method for encoding data into spikes. With rate encoding, we first normalize the data between $0$ and $1$. We then add the time dimension to the data. For example, an image of size $Height \times Width$ becomes three-dimensional matrix of size $N_t \times Height \times Width$, where $N_t$ is the number of time steps we use. Now, instead of an image with pixel values in $[0,1]$ we have $N_t$ images with pixel values of either $0$ or $1$. Although we expand the data, since the new representation is binary, using the sparsity argument we have an even lighter representation of the data in terms of memory usage.

To construct the series of time dependent binary images, we take each pixel in the image and run $N_t$ probabilistic experiments (such as Poisson, Bernoulli, etc.) with expected value equal to the value of the pixel. We get a $N_t$ sized binary vector for each pixel and combining all of them we get the $N_t \times Height \times Width$ binary three-dimensional matrix. In this way we can recover the original image by computing the mean value of each pixel along the time axis. However, reconstruction is not perfect and depends on the number of time steps. With more steps we preserve more information from the original image, but more data is there to process which reduces the efficiency of the SNN.

For function regression, the input data are points on the interval on which the function is defined $\{x_k\}_{k=1}^{N_x}\in [a,b]$, where $N_x$ is the number of discretization points, and $a, b$ are the lower and upper bounds of the interval, respectively. For simplicity, in this work we assume that the points are evenly distributed along the interval. We can use rate encoding here by normalizing the domain using $x\rightarrow \frac{x-a}{b-a}\in [0, 1]$, and create a $N_t \times N_x$ sized two-dimensional matrix as the spiking data using the rate encoding approach.

\subsubsection{Latency encoding}

This is also known as Time-to-First-Spike (TTFS). Following the images example, the idea behind TTFS is that pixels with high intensity hold more information and thus emit a spike earlier. To implement this method we first normalize the data between $0$ and $1$ and quantize this interval with $N_t$ quantiles. A spike is registered for a neuron, in the image case - a pixel, only at the corresponding time to the index of the quantile, where the value of the pixel is. We then flip the entire array so the higher intensity spikes (image pixels with values close to $1$) appear first. This method also preserves the information, and the reconstruction accuracy depends on $N_t$. With TTFS, since every pixel spikes only once, the number of spikes is much lower than with Rate encoding, hence enabling a much more efficient processing.

For functions, a straightforward method to encode the interval is using the identity matrix (ones on the diagonal and zeros elsewhere). The intuition for this is that every point on the interval is equally important, and there are no points of higher intensity. We can think about the point $x=a$ as the lowest intensity point and $x=b$ as the highest, but in both cases the encoding by an identity matrix is in principle the same.

\subsubsection{Lower triangle encoding}

This encoding method was designed for function regression, and the extension to image data is not yet clear. We can think about every point in the evenly distributed interval as a $x=a+n\Delta x$, where $n=0, ..., N_x-1$. In this representation we say that every point has exactly $n+1$ spikes, depending on the position of the point. To implement this, all we need to create is a lower triangular matrix with ones on the lower triangle and zeros elsewhere:

$$
\begin{blockarray}{ccccccc}
    & 0 & \Delta t & 2\Delta t & \cdots & (N_t - 2)\Delta t & (N_t - 1)\Delta t \\
    \begin{block}{c(cccccc)}
        a & 1 & 0 & 0 & \cdots & 0 & 0 \\
        a+\Delta x & 1 & 1 & 0 & \cdots & 0 & 0 \\
        \vdots & \vdots & \vdots & \ddots & \ddots & \vdots & \vdots \\
        b-\Delta x & 1 & 1 & \cdots & 1 & 1 & 0 \\
        b & 1 & 1 & \cdots & 1 & 1 & 1 \\
    \end{block}
\end{blockarray}\quad .
$$

We plan on extending this by changing the location of the spikes per time step. With that we can encode with higher resolution. For example, on the first row we can move the spike between the $N_x$ different points, allowing for a finer discretization of the first sub-interval. We still investigate this option and if found useful it will be included in a future publication.

\subsubsection{Floating point encoding}

Another method to encode data into spikes is by using a binary representation for the data. We fix $N_t$ by the desired floating point precision (usually 32 or 64) and encode each number using the floating point binary representation. As an example, the float 32 representation of the number -25.0313 is (11000001110010000100000000011010). Each 1 in the float 32 representation is a spike.

\subsection{Mathematical formulation of a membrane}\label{membraneDef}

The membrane is a voltage gate for the biological neuron. It is a layer that wraps the neuron and controls the flow of chemicals that go in and out of the neuron. This includes electrically charged ions, setting the voltage. In 1907, Lapicque \cite{lapique1907recherches} conducted an experiment to determine the behavior of the membrane and how it controls the spiking activity of a neuron. The membrane has a threshold that, if crossed, the neuron spikes. We monitor the membrane voltage over time for each neuron, often called the membrane trace, and observe the behavior. The registered spikes of the neurons over time, often called the spikes trace, are the output of the membrane. There are several mathematical models for the behavior of the membrane as mentioned in the introduction. We focus here on two methods for the implementation of the membrane.

\subsubsection{Leaky integrate-and-fire}\label{LIF}

The Leaky Integrate-and-Fire (LIF) model formulates the membrane activity as a Resistor-Capacitor circuit:

\begin{equation}\label{rc_lif}
    C\frac{\partial V(t)}{\partial t} = -\frac{V(t) - V_{rest}}{R} + I(t),
\end{equation}
where $V$ is the membrane voltage, $C$ and $R$ are the capacitance and resistance respectively, $V_{rest}$ is the membrane resting voltage and $I(t)$ is the input current. The solution to this ODE is given by:

\begin{equation}\label{int_lif}
    V(t) = V_{rest} e^{\left( -\frac{t-t_0}{\tau} \right)} + \frac{R}{\tau}\int_{0}^{t-t_0}e^{-\frac{s}{\tau}} I(t-s)ds,
\end{equation}
where $\tau = RC$. To ease the computations we can approximate the solution directly from equation \eqref{rc_lif} using the forward Euler finite difference method, and assume the resting membrane voltage is $0$ to get:

\begin{equation*}
    V(t+\Delta t) = \beta V(t) + I_c(t),
\end{equation*}
where $\beta = 1 - \frac{\Delta t}{\tau}$ and $I_c(t) = \frac{\Delta t}{C}$. This is a rough approximation to the solution of the RC circuit, and lower accuracy in computing the membrane voltage is expected. However, the implementation is much lighter, and allows for more efficient computation.

\begin{figure}
    \centering
    \includegraphics[width=15cm]{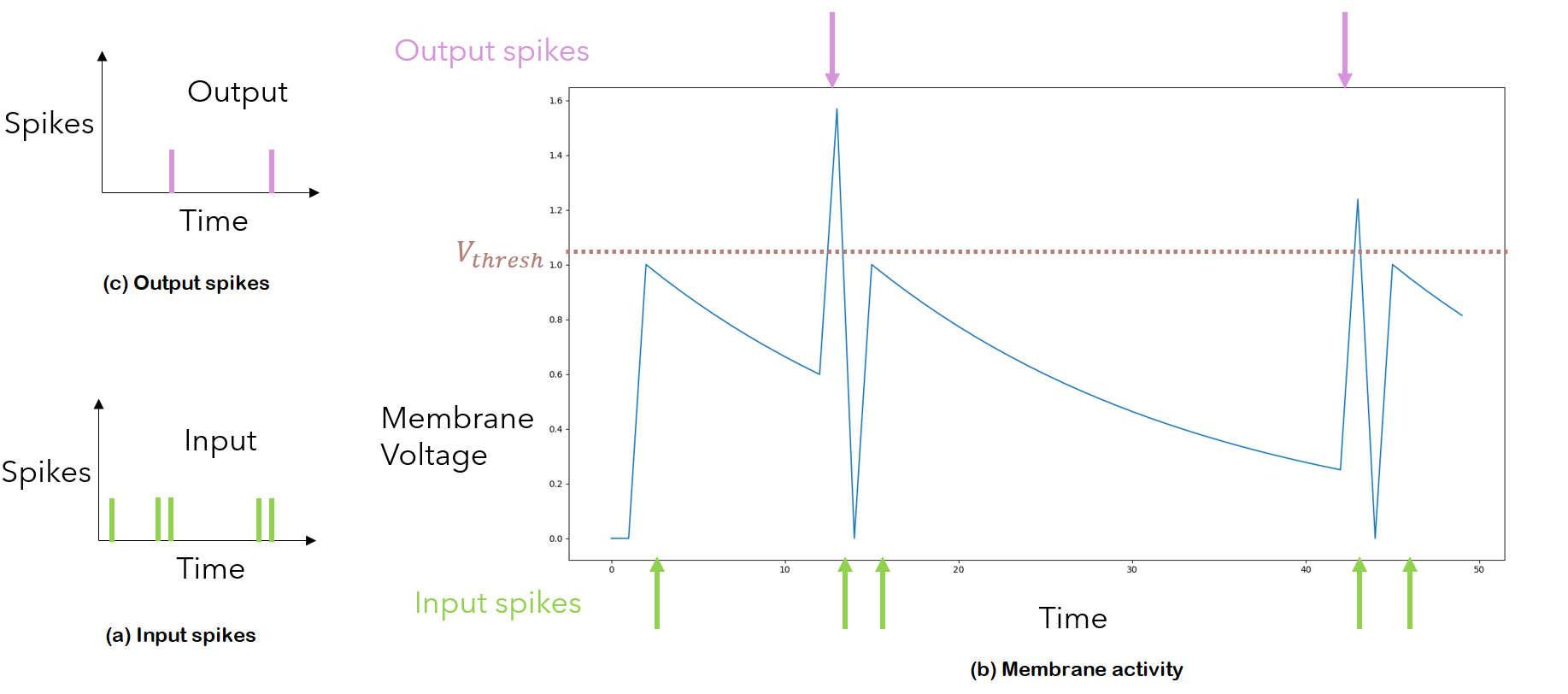}
    \caption{Example of the membrane voltage over time with five input spikes. The reset mechanism is set to resting position without refraction. (a) Presents the input spikes over time. (b) Presents the membrane activity. Notice that when the membrane voltage exceeds the threshold, an output spike is registered and the membrane resets. (c) Presents two registered output spikes over time.}
    \label{fig:lif}
\end{figure}

The membrane has a \textit{threshold} parameter $V_{thresh}$. If the membrane voltage exceeds the threshold, a spike is registered in the neuron; after a spike the membrane \textit{resets}. There exist various methods for resetting the membrane; two common ones are:
\begin{itemize}
    \item \textbf{Reset to resting position}: After a spike the membrane voltage resets to $V_{rest}$.
    \item \textbf{Reset by voltage subtraction}: After a spike the membrane voltage drops by a constant voltage, usually the threshold $V_{thresh}$.
\end{itemize}
In addition, after a spike the neuron enters a refractory state, where it does not spike again for a few time steps. One way to achieve that is to reset the membrane voltage by, for example, $2V_{thresh}$ so incoming currents do not immediately create a spike. Another method is to higher the threshold $V_{thresh}$ after a spike is registered. An example of membrane voltage over time for a set of input spikes is given in figure \ref{fig:lif}. We observe the leaky behavior after the first input spike, and the reset after exceeding the threshold.

\subsubsection{Multilayer-Perceptron membrane}\label{MLP_membrane}

Another way to model a membrane is to learn the membrane dynamics using a Multilayer-Perceptron (MLP). MLPs are fully-connected feed-forward ANNs. We address the membrane dynamics as a function, and train a MLP to approximate this function. We emphasize that the dynamics include the membrane threshold and the reset mechanism (not just the solution of the ODE). The inputs to the MLP are the input spikes, and the outputs are the output spikes, similar to the membrane. Since the MLP is an ANN, the accuracy of producing (in this case, inferring) the output spikes depends on the quality of the training and the input data. The former can be enhanced with the choice of hyper-parameters. The latter is more interesting - one can use more accurate learning models to create the data for training the MLP. For instance, solving the ODE directly by approximating the integral (and not using the forward Euler method) produces more accurate input data, and thus a more accurate  MLP-membrane. We experiment with several learning models and the results are presented in section \ref{results}.

The main advantage of using an MLP membrane is that inference is not expensive compared to the other methods. When using the MLP to compute the output spikes, the input data is fed to the MLP and goes through a few matrix multiplications. Other methods require more complex operations. For example, approximating the integral solution of the ODE requires time marching (loop over the number of time steps), and computing the integral using a (usually costly) numerical integrator. When comparing the different membrane implementations, in addition to the accuracy we also compare the time it takes to produce the solution.

\begin{figure}
    \centering
    \includegraphics[width=15cm]{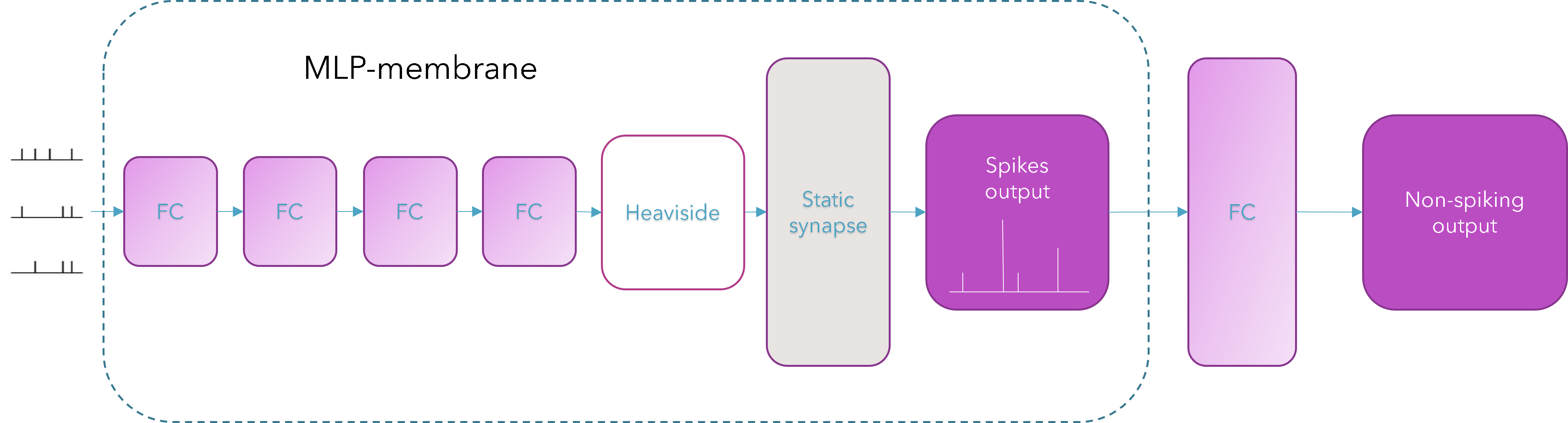}
    \caption{A MLP-membrane based SNN. We infer the output of the LIF using the illustrated architecture. The first four boxes on the left are trainable Fully-Connected layers (FC), followed by a Heaviside activation to create the spikes. The MLP-membrane (bounded using a dashed box) is trained offline and then replaces a LIF layer in an SNN. In this illustration we consider a SNN with one LIF layer and one dense layer, learning to infer the non-spiking output. Note that in this case we pass the output of the Heaviside function in the MLP-membrane as is, without multiplying with any weights, hence the usage of a static synapse in this work.}
    \label{fig:MLP_membrane}
\end{figure}

To integrate the MLP-membrane into a SNN, the architecture we use includes four fully-connected layers and a Heaviside operation at the end, to convert the outputs of the MLP back into spikes. Each of the four dense layers has 100 neurons. We train the MLP-membrane separately (offline training) and substitute the LIF layer of the SNN with the trained MLP membrane layer. An illustration of the architecture used for the MLP-membrane is presented in figure \ref{fig:MLP_membrane}. In this illustration we show the four fully-connected layers of the MLP followed by the Heaviside function. We do not use a learning rule to procude the output spikes (referred to as static synapse, which is explained in the next section). The four dense layer, the Heaviside operator and the output spikes define MLP-membrane and can be substituted with a LIF layer. To investigate the performance of the MLP-membrane we add a dense layer at the end, learning of the classification or regression task, making this a SNN.

\subsection{Mathematical formulation of a synapse}\label{synapse}

The synapse is the component in the brain where the learning part happens. The synapse enables electrical or chemical signals to transfer between one neuron to another. In terms of implementation, the synapse controls the connection between the input and output neurons. The synapse has weights $W_{i,j}$, where higher $W_{i,j}$ means stronger connection between input neuron $i$ and output neuron $j$. These weights multiply the input current $I(t)$ in equation \eqref{rc_lif}. To train the synapse means apply a Hebbian learning rule to determine the weight updates after each sample is propagated through the network. We discuss two approaches for updating the weights. Note that for some of the experiments made through this work we did not use a trainable synapse and replaced it with a simple identity mapping (all the weights between the neurons are set to 1). This is referred to as a \textit{static synapse}.

\subsubsection{Spiking Time Dependent Plasticity (STDP)}\label{STDP}

In STDP the weight updates are determined by the temporal difference between the spikes. We split the set of spikes into pre-synaptic spikes (input spikes, before going through the synapse) and post-synaptic spikes (output spikes). The split is based on the temporal order of the spikes in the spikes train, activating the different neurons in the brain at a sequential order.

The STDP rule says that if a pre-synaptic neuron spiked before the post-synaptic neuron, the connection between these neurons is increased, otherwise known as potentiation. If the post-synaptic neuron spiked before, the connection is decreased, otherwise known as depression. The weight change can be computed by the following formula \cite{stimberg2019brian}:

\begin{equation}\label{stdp_curve}
    \Delta W_{i,j}(\Delta t_{i,j}) =
    \begin{cases}
        A_{pre}e^{-\frac{\Delta t_{i,j}}{\tau}} &, \Delta t_{i,j} \geq 0 \\
        -A_{post}e^{\frac{\Delta t_{i,j}}{\tau}} &, \Delta t_{i,j} < 0,
    \end{cases}
\end{equation}
where $\Delta t_{i,j}$ is the time difference between the pre-synaptic spike and the post-synaptic spike and $A_{pre}$ and $A_{post}$ are constants controlling the learning rate. Using equation \eqref{stdp_curve}, also known as the STDP curve, we compute the weight changes, and learn the synapse weights. It is important to mention that this rule can be applied in the forward propagation. It does not require back-propagation, and no gradients are involved. Also, the weight updates can be computed on different compute units for different input samples and then summed back to compute the total weights, hence enabling parallel computing.

A drawback of the STDP method is that it is unstable. Using STDP as described here, the synaptic strengths grow or shrink indefinitely unless limits are imposed. Specifically, given the STDP curve used here, the weight changes are relatively large, creating large weights after a small set of samples fed to training. To overcome this, we chose the parameters $A_{pre}$ and $A_{post}$ so that smaller changes occur, granting us small synapse values for the number of samples used in the training phase. This can be handled better using limits for the synaptic weights for example, as shown in \cite{babadi2010intrinsic}. Still, we did not manage to properly create a stable STDP based synapse for function regression, and this topic is being investigated. STDP is the most popular implementation of a Hebbian learning rule for the synapse, and although we do not use it directly in the current work, it provides a good intuition for the following sections.

\subsubsection{Fully connected weights}

A popular method to learn the synaptic connections between the input and output neurons is using a Fully Connected (FC or dense) layer. In this way, the weights are learned by back-propagation. With the advanced methods of DL optimization, the dense network can train fast and produce relatively accurate results. Therefore, this method has been adopted by many authors. We present results using this method as well.

\subsection{Spikes decoding}

After the data passed through the membrane and the synapse, we get the output spikes. For classification and regression tasks we want to decode these spikes into data. For MNIST classification, the number of neurons in the last layer is set to 10 according to the 10 different MNIST classes ($0-9$ digits). However, neuron number 1 does not necessarily represent the digit $0$. We need to connect the output spikes to the digit they represent. We can achieve this by using a Support Vector Machine (SVM). Another way is to use a dense layer  as the last hidden layer. This means we back-propagate only on the last layer, which is not computationally expensive.

For function regression we can either use a dense layer at the end to decode the spiking outputs back to numbers, or we can use floating point decoding, which is the opposite of the aforementioned floating point encoding. The latter means that the number of output neurons is the precision we want (usually 32 or 64), and for this to work we need to make sure that the output neurons correspond to the positions in the floating point representation. One way to do that is to use a Hebbian learning rule that can handle supervised tasks. In other words, the order of the output neurons is important for reconstructing the output of the network back into a number. This approach is currently being investigated, and instead throughout this paper we use a dense layer to decode the spiking outputs.

\section{Spiking DeepONet}

\begin{figure}
    \centering
    \includegraphics[width=15cm]{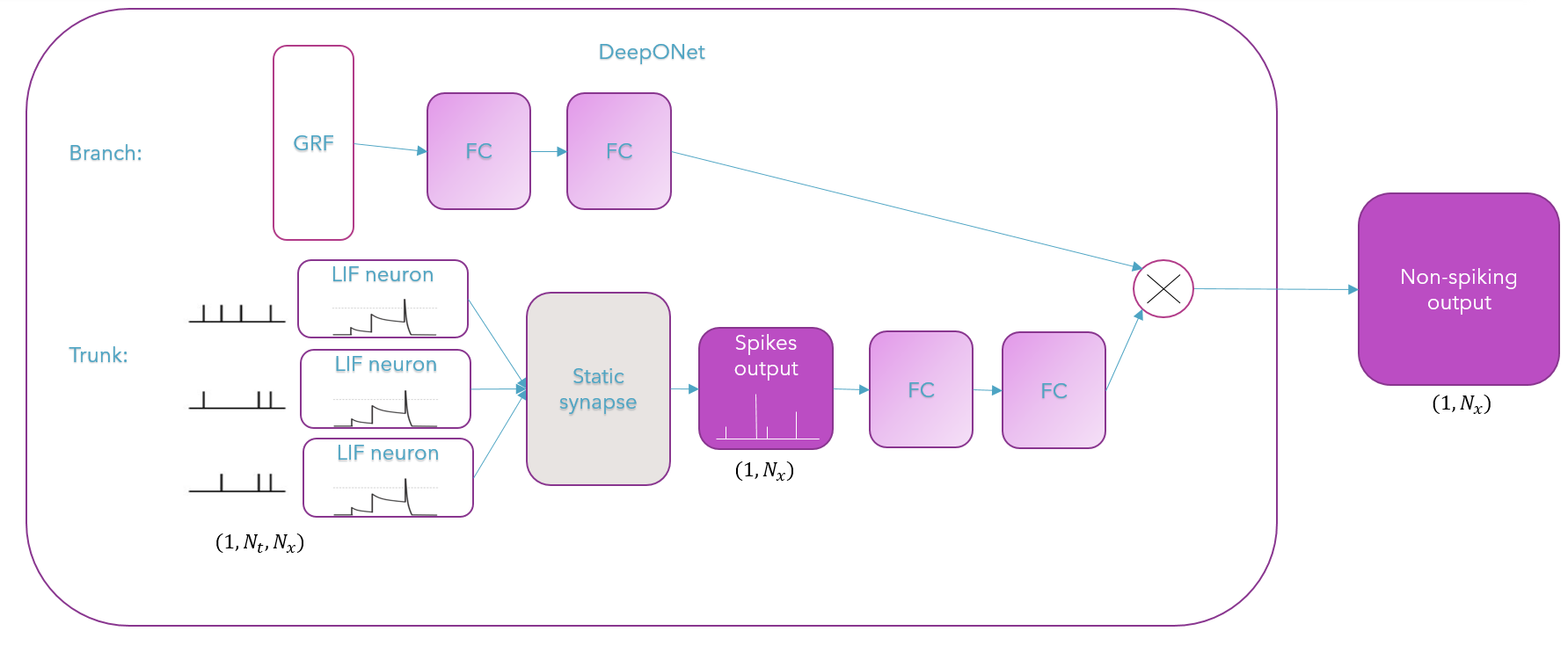}
    \caption{The structure of the spiking DeepONet. In this design the trunk is a SNN while the branch is a standard FNN. The inputs to the branch are arbitrary functions generated from GRF while the inputs to the trunk are spiking data. The outputs of the branch and the trunk are multiplied to produce the non-spiking output.}
    \label{fig:spiking_deeponet}
\end{figure}

In this section we discuss the proposed method of using DeepONets in a SNN. We use a un-stacked DeepONet with a single branch and a single trunk. To investigate the ability of DeepONet to handle spiking data, we first use it for classification. In this case, the branch is taken as the SNN. Then, for the regression task, the trunk inputs are spiking data while the branch inputs are Gaussian Random Field (GRF) arbitrary functions. The details and hyper-parameters defining the branch network and the trunk network are given in section \ref{results}, where we discuss the setups we use for testing the methods. An illustration of the DeepONet framework for regression is given in Figure \ref{fig:spiking_deeponet}. In this architecture the trunk is a SNN consisting of a membrane, followed by a static synapse (the synapse weights are constant), and the learning part of the trunk is done by 2 fully-connected layers after the spikes trace has been computed. The output of the DeepONet is non-spiking data, such as the MNIST digits (classification) or function values (regression).

\subsection{MNIST classification}

As a first experiment, it is interesting to check how the spiking DeepONet framework performs compared to popular benchmarks. Most studies of SNNs use the MNIST data to train and infer. We have two benchmarks in this comparison: 1) State-of-the-art ANNs for MNIST classification, and 2) SNNs used for MNIST classification. The SNNs usually have lower accuracy on this data compared to the ANNs, and we want to see how the spiking DeepONet performs compared to the other available competing methods. Using the DeepONet for this task is challenging since the architecture is not designed for classification. We propose a method for using DeepONet with spiking inputs to classify the MNIST digits, which is different from the one we propose for function regression.

To create the data for the spiking DeepONet, we first use rate encoding to convert the MNIST data into spiking data. The inputs to the branch are the encoded MNIST spiking data. The inputs to the trunk are digits $0-9$. The output is a $1$ if the input digit for the trunk matches the encoded digit passed into the branch, and $0$ otherwise. This process can be interpreted as learning an operator that maps from spiking input digits (interpreted as a function) to a Kronecker delta whose first argument is spiking input (representing a digit) passed into the branch, and whose second argument is the digit passed into the trunk. The branch uses a LIF layer followed by dense layers, and the trunk uses dense layers. More information about the implementation such as the number of neurons, the DeepONet hyper-parameters and the chosen optimization technique are given in section \ref{results}.

\subsection{Function regression}

For function regression, the usage of DeepONets is completely different. Since DeepONets are designed to learn operators, function regression is a more natural task. However, the input spikes are a challenge for the DeepONet framework. The underlying assumption is that the process in the brain that involves the membrane (including the thresholds, the resets, etc.), the synapse, and so on, can be formulated as one large operator that DeepONet can learn. We propose a method to do regression with spike-encoded inputs, and show that DeepONet can actually learn this unknown operator.

In this case, we generate a set of functions using GRFs, and encode the input interval into spikes using one of the methods described in section \ref{encoding} (we compare the performance of the spiking DeepONet with different encoding methods in section \ref{results}). We use the set of random functions as inputs to the branch and the spiking data as inputs to the trunk. The branch uses dense layers while the trunk uses a LIF layer followed by a static synapse (no learning rule applied for the synapse in this case) and two dense layers that are trained using back-propagation. An illustration of this architecture is given in figure \ref{fig:spiking_deeponet}.

\section{Results}\label{results}

We conduct multiple numerical experiments using conventional hardware (CPU and GPU, no neuromorphic processors). The spike encoding method used changes between the tests, and we state which encoding method has been used for each test. The LIF model for the membrane used in the tests is the one presented in section \ref{LIF} with membrane resting voltage $V_{rest} = 0$, membrane threshold $V_{thresh} = 1$, and $\beta = 0.95$. For the synapse we either use a static synapse (constant weights) or fully-connected weights, and in each test we mention which synapse implementation is used; same for the spike encoding and decoding methods.

\subsection{Classification results}

Although the main result of this paper regards regression, we first want to test the performance of the proposed method for classification. Most of the reference studies are studying classification tasks, and we can compare to other state-of-the-art methods (ANNs and SNNs). We use the MNIST data set that contains 70,000 images (60,000 train and 10,000 test) of black and white handwritten digits of size $28\times 28$ pixels. This data set is a very popular benchmark for SNNs. We compare the results to a state-of-the-art ANN method developed by An \textit{el atl.} \cite{an2020ensemble}, and to the state-of-the-art SNN method developed by Diehl \textit{et al.} \cite{diehl2015unsupervised}. We also compare an implementation of DeepONet (for non-spiking data). The branch of this DeepONet is a convolutional neural network with three convolution layers with 16, 32 and 32 filters of kernel size (5, 5) with ReLU activation, followed by two dense layers of size 50 each. The trunk is a FNN with 3 dense layers of size 50. The input to the branch are the MNIST images and the input to the trunk is the MNIST digit.

In addition, we add to the comparison a method proposed by Youngeun \textit{et al.} \cite{kim2022rate}, discussing direct encoding. In their paper the authors show that direct encoding is more accurate than rate encoding. We use a similar SNN architecture as proposed in that paper - two dense layers of size 800 and 10, respectively, each followed by a LIF membrane. We use reset to zero (instead of the reset by subtraction that the authors used), and achieved even better results. We experiment with different number of time steps varying between $6$ and $50$, and the results are only marginally different as expected. 

The proposed method we benchmark is the proposed spiking DeepONet used for MNIST classification. The DeepONet branch has a LIF layer for the membrane with $784$ input neurons (the flattened $28\times28$ MNIST images), followed by four fully-connected layers of sizes $512, 250, 50, 1$, respectively, with ReLU activation. The trunk has two fully-connected layers of size $50$ with ReLU activations. The results are given in table \ref{tab:mnist}. All methods use the Adam optimizer \cite{kingma2014adam}. We observe that the proposed spiking DeepONet is competitive with the state-of-the-art models, especially when comparing to the SNN benchmark.

\begin{table}[ht!]
 \caption{MNIST classification comparison table. The spiking DeepONet here has a SNN as the branch and standard FNN trunk.}
  \centering
  \begin{tabular}{lll}
    \toprule
    Method & Accuracy \\
    \midrule
    ANN (An \textit{et al.}, 2020) & $99.91$\% \\
    SNN (Diehl \textit{et al.}, 2015) & $99.1$\% \\
    DeepONet & $99.3$\% \\
    Direct encoding & $98.3$\% \\
    Spiking DeepONet & $99.08$\% \\
    \bottomrule
  \end{tabular}
  \label{tab:mnist}
\end{table}

\subsection{Regression results}

\subsubsection{Naive approach results}

The first test we discuss is a naive method for function regression with spiking input data. Given data of a function $y(x)$, we encode the inputs $x$ into spikes using one of the spike encoding methods mentioned in section \ref{encoding}. The goal is to train a network to fit spike encoded inputs to the function values $y(x)$. The naive approach uses a simple architecture and does not involve a DeepONet structure, which means that in this case we are not learning an operator, but directly learn the function itself. The challenge here is that the inputs are spikes, and not decimal numbers as in most ANNs. This is a feasibility study - we first show that it is possible to perform function regression in a spiking regime, and later improve the performance by using the DeepONet. The input to the proposed network is a spike-encoded interval as described in section \ref{encoding} with $N_t = 50$ time steps. The interval is $[0, 1]$ and we select $N_x = 100$ points for the spatial discretization. In this test we use a latency encoder (identity matrix). The proposed architecture is a LIF layer with input sample size $N_t\times N_x$. The output of the LIF layer is a spikes trace, explained in section \ref{membraneDef}. We sum the number of spikes at the output of the membrane (sum the spikes trace over time), so the output of the LIF is the total number of spikes per neuron. Then, we add a dense layer of size $N_x$ with ReLU activation. We have a single sample, which is the function we are trying to approximate, that includes all the points and corresponding function values ($\{x_n\}_{n=0}^{N_{x}-1}\rightarrow\{y(x_n)\}_{n=0}^{N_{x}-1}$). In other words, network weights are trained to fit an input vector of size $N_x$ to an output vector of the same size. The values of the function on the entire interval are predicted at once. We perform separate training for each function. We mention that we did try predicting point-by-point (each point is a sample, $\{x_n\rightarrow y(x_n)\}_{n=0}^{N_{x}-1}$), but although the network was able to predict the approximated solution, it was much worse than the proposed inference of $N_x$ points altogether.

\begin{figure}[b!]
     \centering
     \begin{subfigure}[b]{0.3\textwidth}
         \centering
         \includegraphics[width=\textwidth]{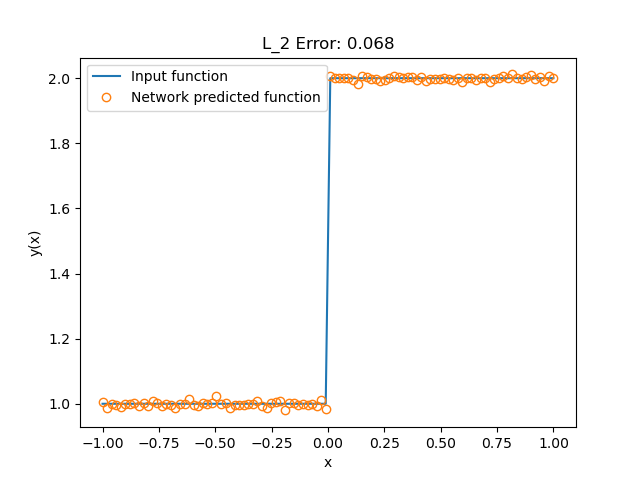}
         \caption{Discontinuity.}
     \end{subfigure}
     \hfill
     \begin{subfigure}[b]{0.3\textwidth}
         \centering
         \includegraphics[width=\textwidth]{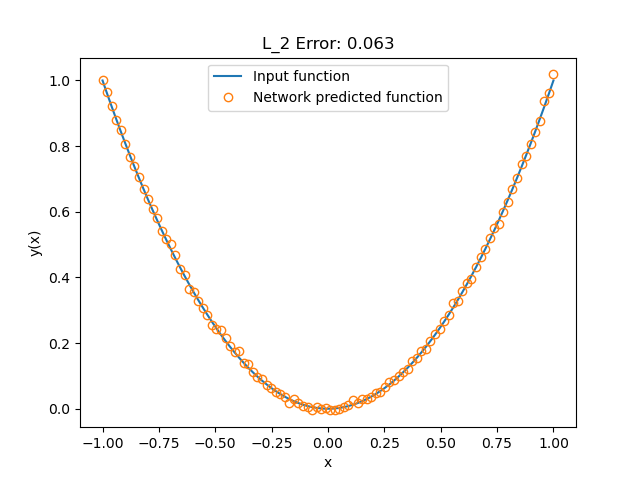}
         \caption{Square.}
     \end{subfigure}
     \hfill
     \begin{subfigure}[b]{0.3\textwidth}
         \centering
         \includegraphics[width=\textwidth]{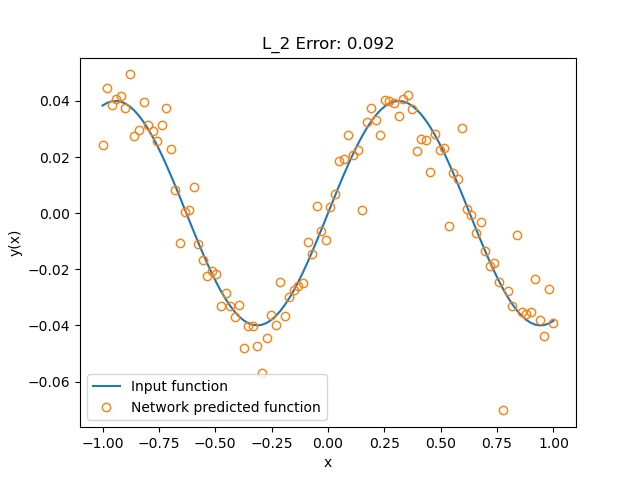}
         \caption{Sine.}
     \end{subfigure}
     \begin{subfigure}[b]{0.3\textwidth}
         \centering
         \includegraphics[width=\textwidth]{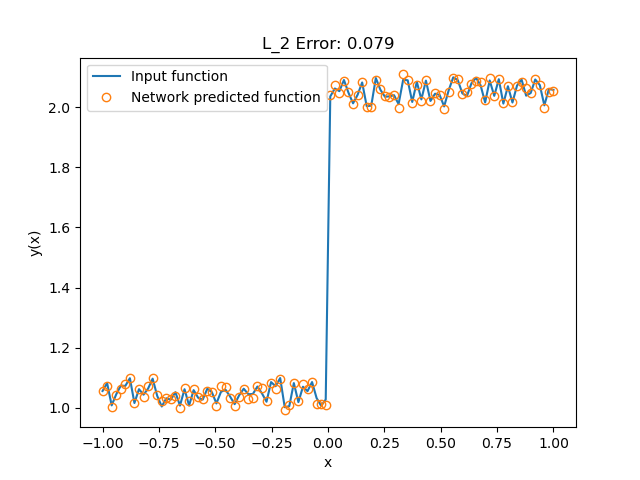}
         \caption{Noisy discontinuity.}
     \end{subfigure}
     \hfill
     \begin{subfigure}[b]{0.3\textwidth}
         \centering
         \includegraphics[width=\textwidth]{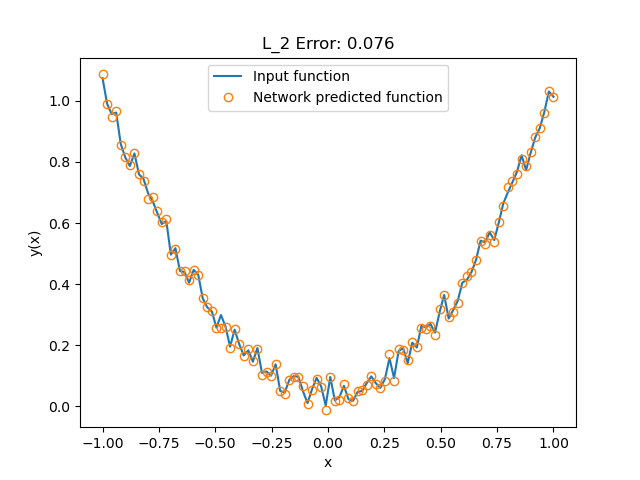}
         \caption{Noisy square.}
     \end{subfigure}
     \hfill
     \begin{subfigure}[b]{0.3\textwidth}
         \centering
         \includegraphics[width=\textwidth]{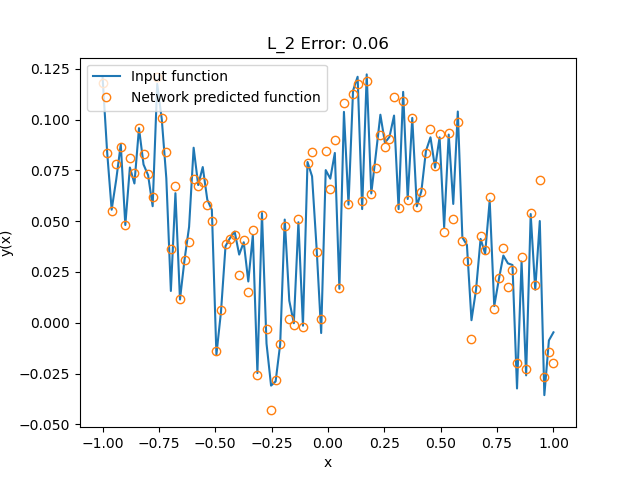}
         \caption{Noisy sine.}
     \end{subfigure}
        \caption{Feasibility study of function regression using the naive SNN implementation for different functions described in \ref{tab:naive_snn_results}.}
        \label{fig:naive_snn_reg}
\end{figure}

We demonstrate the success of the naive method in Figure \ref{fig:naive_snn_reg}. Each pair of images (top and bottom) shows the input function in a solid line and the network prediction using circles. The leftmost plots show the network prediction for an input discontinuity function, the middle for a square function, and the rightmost plots show a simple sinusoidal ODE. In the second row, the functions are perturbed with $0.1$ additive normally distributed noise, whereas in the first row no noise has been added. We compare the true and predicted results via the $L_2$-norm given by $\sum_{j=1}^{N_x}\left(y_{pred}(x_j) - y_{true}(x_j)\right)^2$. The results are summarized in table \ref{tab:naive_snn_results}. Interestingly, for the sinusoidal example the SNN performed better in the noisy experiment than in the noise-less one, in terms of the $L_2$ error.

\begin{table}[t]
 \caption{Naive SNN regression results.}
  \centering
  \begin{tabular}{lll}
    \toprule
    Function & Noise-less data, $L_2$-norm error & Noisy data, $L_2$-norm error \\
    \midrule
    $y=\begin{cases} x = 1 & ,-1\leq x \leq 0 \\ x = 2 & ,0 < x \leq 1 \end{cases}$ & 0.068 & 0.079 \\ \\
    $y=x^2$ & 0.063 & 0.076 \\ \\
    $y_{xx}(x)=\sin(kx)$ & 0.092 & 0.06 \\
    \bottomrule
  \end{tabular}
  \label{tab:naive_snn_results}
\end{table}

\subsubsection{MLP-membrane results}

In this experiment we substitute the LIF layer with a MLP-membrane trained to learn the membrane behavior (on spiking data). The MLP-membrane is trained separately (offline training) on MNIST data to fit input spikes to output spikes. To generate the data we use the MNIST set and encode it using rate encoding as described in section \ref{encoding}. The reason for that is the MNIST data, after using rate encoding, has a rich number of spikes which results high membrane activity. This enables the MLP-membrane to efficiently learn the membrane dynamics. We use the forward Euler implementation for producing the LIF outputs and create the labels for training the MLP-membrane model. The LIF-membrane architecture we use is four dense layers of sizes [100, 100, 100, 100] followed by a Heaviside operation. After the training is done, we fix the weights of the MLP-membrane and use it for regression. We create a SNN using one LIF layer and one dense layer, and train it to approximating a Ricker wavelet in two dimensions given by:
\begin{equation*}
    \psi(x, y) = \frac{1}{\pi \sigma ^4}\left(1 - \frac{1}{2}\left(\frac{x^2 + y^2}{\sigma^2}\right)\right)e^{-\frac{x^2 + y^2}{2\sigma^2}},
\end{equation*}
with $\sigma=0.4$. For training the SNN to infer the Ricker wavelet, we encode the input $x$ and $y$ and concatenate them together. The output then is the number $\psi(x, y)$. This means that in this case we predict point-to-value $\{(x_n, y_n)\rightarrow \psi(x_n, y_n)\}_{n=0}^{N_{points}-1}$, where $N_{points} = 10,000$ uniformly distributed on a grid in $[-1, 1]\times[-1, 1]$ (100 points in each direction). We compare the MLP-membrane to a standard LIF based membrane, and also to a ``No membrane'', where in the spiking activation layer we do not apply neither LIF nor the trained MLP-membrane, and just use the dense layer following the membrane for the training. We only substitute the LIF part, the following dense layer is the same for all three methods. The results are given in table \ref{tab:membrane_res_tab}. We also monitor the time it takes to process a single sample from the training set. We see that the approximate LIF produces the most accurate results, while the MLP-membrane achieves relatively accurate results with much faster time per sample.

\begin{table}[b!]
 \caption{Function regression by different membrane models.}
 \hspace*{-0.2cm}
  \centering
  \begin{tabular}{lllll}
    \toprule
    Encoding & Mean $L_2$-norm error & Median $L_2$-norm error & Std of $L_2$-norm error & Time per sample(ms) \\
    \midrule
    No membrane & 0.036 & 0.036 & 0.001 & 0.631 \\
    MLP-membrane & 0.020 & 0.020 & 0.001 & 0.933 \\
    LIF & 0.015 & 0.015 & 0.001 & 3.107 \\
    \bottomrule
  \end{tabular}
  \label{tab:membrane_res_tab}
\end{table}

\subsubsection{Spiking DeepONet results}

\begin{figure}
     \centering
     \begin{subfigure}[b]{1\textwidth}
         \centering
         \includegraphics[width=\textwidth]{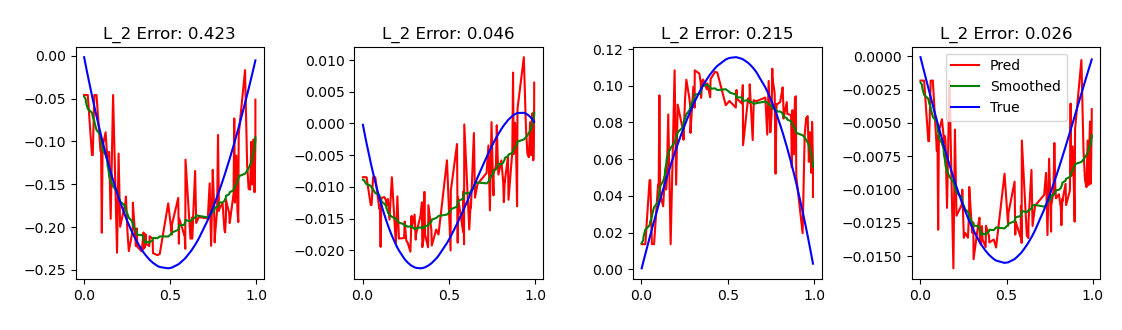}
         \caption{Rate encoding.}
     \end{subfigure}
     \hfill
     \begin{subfigure}[b]{1\textwidth}
         \centering
         \includegraphics[width=\textwidth]{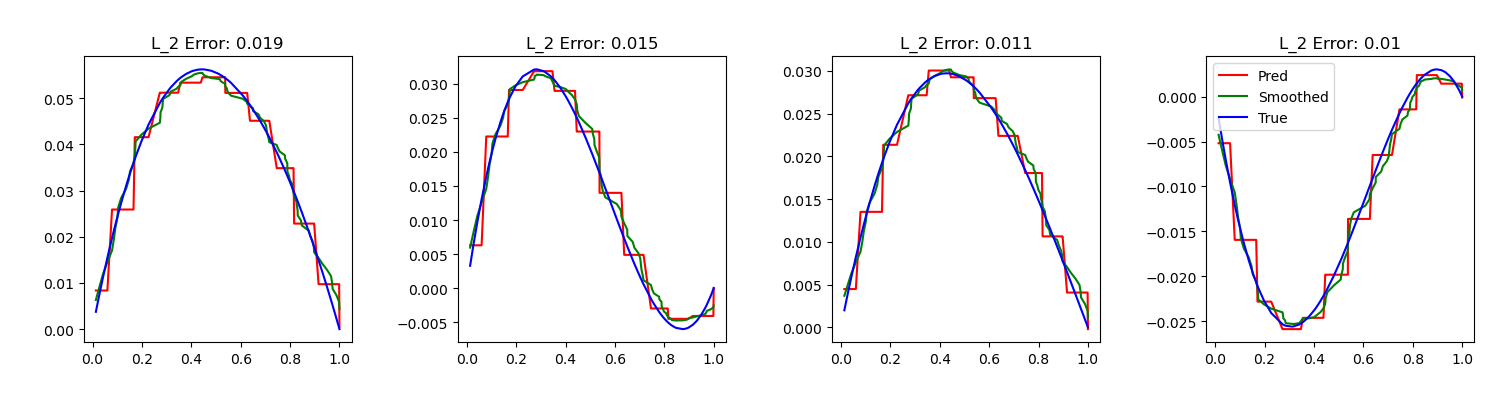}
         \caption{Lower triangular encoding.}
     \end{subfigure}
     \begin{subfigure}[b]{1\textwidth}
         \centering
         \includegraphics[width=\textwidth]{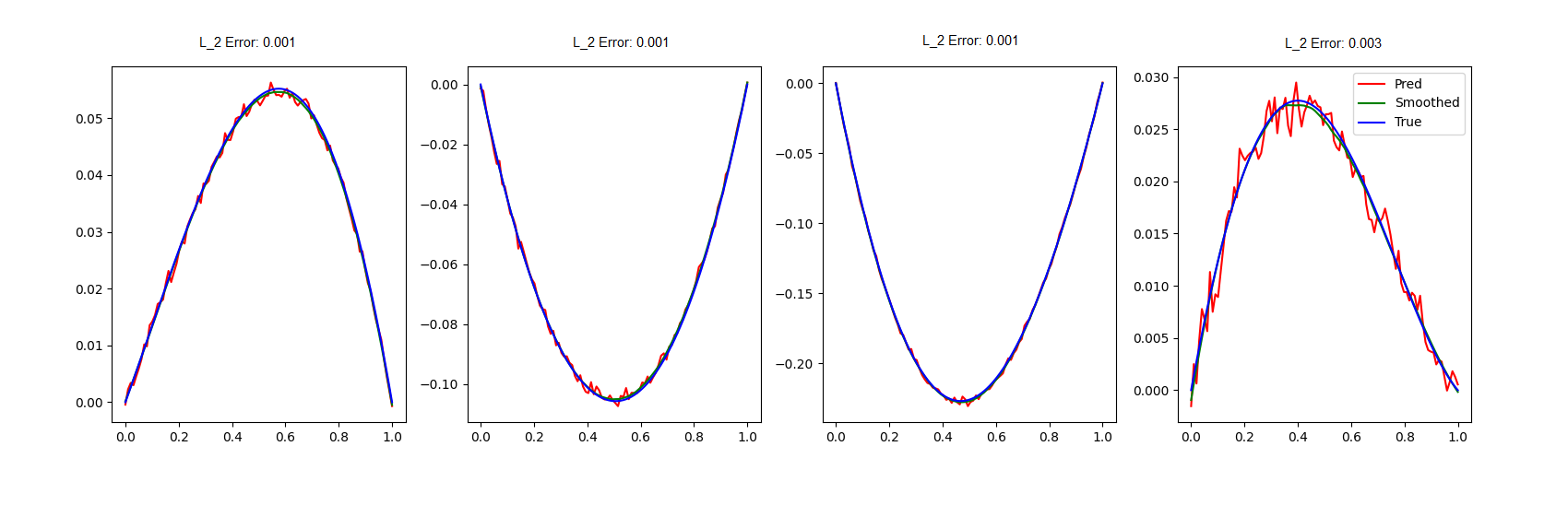}
         \caption{Lower triangular encoding with 10x time steps.}
     \end{subfigure}
     \begin{subfigure}[b]{1\textwidth}
         \centering
         \includegraphics[width=\textwidth]{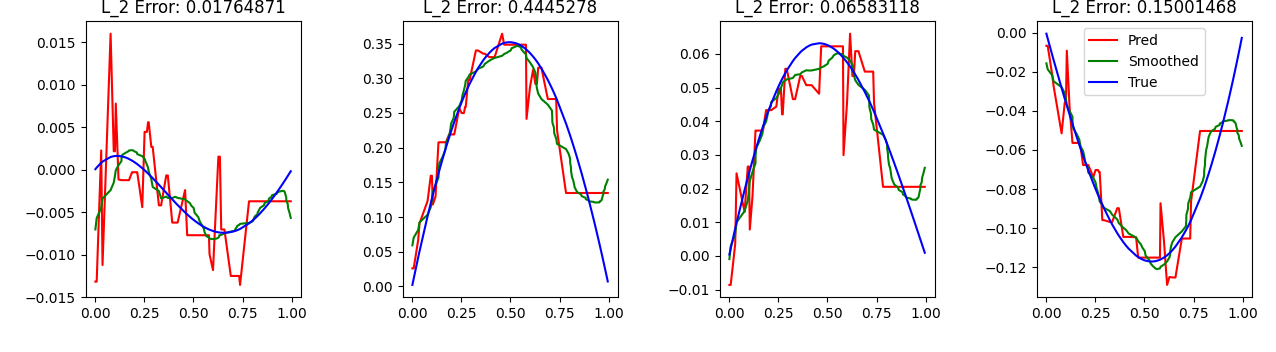}
         \caption{Direct encoding.}
     \end{subfigure}
        \caption{Function regression using spiking DeepONet for different encoding methods for the trunk input. The proposed triangular b-c) encoding outperforms the existing encoding methods.}
        \label{fig:spiking_deeponet_results}
\end{figure}

To show the performance of the spiking DeepONet we first generate $1,000$ GRF samples by solving the one dimensional Poisson equation for different right hand sides (we refer the readers to the DeepONet paper \cite{lu2021learning} for further information about training a DeepONet). We use a length scale of 1 for the radial basis functions kernel. We split this data-set so that $N_{train} = 800$ samples are used for training and $N_{test} = 200$ samples are used for testing. The branch uses two dense layers of size $30$, and the trunk uses a LIF layer followed by two dense layers of size $30$. The input to the branch are the GRF functions, the input to the trunk are the spike encoded spatial points and the outputs of the DeepONet are the function values. We train the DeepONet to infer the function values from the input encoded data. The number of time steps is set to $N_t = 50$, and the spatial discretization is set to $N_x = 100$. We test the method with rate, floating point, lower triangular and direct encoding as described in section \ref{encoding}. We use a Savitzky-Golay smoothing filter \cite{press1990savitzky} with window size of 51 and polynomial order of 3, and compare the true solution to the smoothed solution via the $L_2$-norm. We observe that the simple smoother we use produces more accurate results and justifies the smoothing operation. Figure \ref{fig:spiking_deeponet_results} presents an example of inference of four different randomly generated solutions of the Poisson equation using the spiking DeepONet. In the upper four images, rate encoding has been used. In the second row we show the results using the lower triangular encoding. In the third, a finer usage of the lower triangular encoder (10 $\times$ $N_t$). In the lower row we used the direct encoding. We observe the stair-casing behavior when using the lower triangular method which comes from the encoding shape. By using a larger number of time steps the stair-case solution is finer and more accurate, but the computation time increases. In table \ref{tab:spiking_deeponet_res_tab} we present the $L_2$-norm error in terms of the mean and median values (over the entire training set). We see that the lower triangular method triumphs in this case, performing better than the alternative methods.

\begin{table}[ht!]
 \caption{Spiking DeepONet results for solving the one dimensional Poisson equation using different encoding methods.}
 \hspace*{-0.2cm}
  \centering
  \begin{tabular}{llll}
    \toprule
    Encoding & Mean $L_2$-norm error & Median $L_2$-norm error & Std of $L_2$-norm error \\
    \midrule
    Rate & 0.229 & 0.182 & 0.174 \\
    Floating point & 0.692 & 0.659 & 0.451 \\
    Lower triangular & 0.047 & 0.043 & 0.032 \\
    \textbf{Lower triangular and 10x time steps} & \textbf{0.005} & \textbf{0.004} & \textbf{0.003} \\
    Direct encoding & 0.135 & 0.121 & 0.067 \\
    \bottomrule
  \end{tabular}
  \label{tab:spiking_deeponet_res_tab}
\end{table}

\section{Conclusion}

In this work we discussed using SNNs for performing scientific computations. We first introduced the key elements of constructing a SNN and their biological plausibility. Then we proposed a method to perform regression using SNNs, and conducted a feasibility study showing that it can be done. Then, we showed that the expensive computation of the membrane dynamics can be replaced by a highly efficient MLP network without losing much accuracy. Finally, we introduced a spiking DeepONet and used it to learn the spiking behavior. We first showed how it fairs against existing methods, which are classification based. Then we extended it to regression and showed even better performance on inferring the solution of differential equations. These results show promise for efficient function regression using sparse spiking data, and in the future, using neuromorphic hardware we will be able to accelerate these computations much further.


\bibliographystyle{unsrt}  
\bibliography{references}

\end{document}